# CST-YOLO: A NOVEL METHOD FOR BLOOD CELL DETECTION BASED ON IMPROVED YOLOV7 AND CNN-SWIN TRANSFORMER

*Ming Kang, Chee-Ming Ting*, Fung Fung Ting, Raphaël C.-W. Phan*

School of Information Technology, Monash University, Malaysia Campus

## ABSTRACT

Blood cell detection is a typical small-scale object detection problem in computer vision. In this paper, we propose a CST-YOLO model for blood cell detection based on YOLOv7 architecture and enhance it with the CNN-Swin Transformer (CST), which is a new attempt at CNN-Transformer fusion. We also introduce three other useful modules: Weighted Efficient Layer Aggregation Networks (W-ELAN), Multiscale Channel Split (MCS), and Concatenate Convolutional Layers (CatConv) in our CST-YOLO to improve small-scale object detection precision. Experimental results show that the proposed CST-YOLO achieves 92.7%, 95.6%, and 91.1% mAP@0.5, respectively, on three blood cell datasets, outperforming state-of-the-art object detectors, e.g., RT-DETR, YOLOv5, and YOLOv7. Our code is available at `https://github.com/mkang315/CST-YOLO`.

## 1. INTRODUCTION

Automated blood cell detection refers to recognizing different types of blood cells, including red blood cells (RBCs), white blood cells (WBCs), platelets, etc., in microscopic images. It is a crucial process for accurate blood cell count in pathology labs used for the diagnosis and treatment of different diseases. The main challenge in blood cell detection is that blood cells are small-scale objects for which traditional object detectors can only achieve sub-optimal performance. Faster Region-based Convolutional Neural Network (Faster R-CNN) [1, 2, 3], YOLOv3 [4], and YOLOv5 [3, 5] have achieved the best precision so far in automated detection of blood cells, where YOLOv5 performs better than Faster R-CNN [3].

YOLOv7 [6] is a cutting-edge object detector that surpasses YOLOv5 [7] and many other object detectors in speed and accuracy. The basic components of YOLOv7 networks that play a major role are ConvBNSiLU (or CBS), MPConv, Extended Efficient Layer Aggregation Networks (E-ELAN), and Spatial Pyramid Pooling & Cross Stage Partial Network plus CBS (SPPCSPC). CBS is a Convolutional (Conv) layer followed by a Batch Normalization (BN) layer, and then finally a Sigmoid-weighted Linear Unit (SiLU) [8] as an activation function, which was firstly used by EfficientNet [9].

CBS replaces ConvBNLeakyReLU (or CBL) in YOLOv5 v4.0 and newer versions, and was also adopted by PP-YOLOv2 [10], YOLOX [11], PP-YOLOE [12] and YOLOv8 [13]. MPConv connects two MaxPool + CBS layer branches and a CBS + CBS layer. E-ELAN controls the connection paths of different lengths, enabling the network to learn and convert effectively. Inspired by SPP [14] and CSPNet [15], SPPCSPC makes the head network suitable for multi-scale input and achieves a fusion of different levels of features. Recently, [16] proposed Swin Transformer, a hierarchical Transformer designed for computer vision tasks. It implements self-attention within local windows and establishes long-range cross-window dependency by shifted windowing scheme. A hybrid network of convolutional neural network (CNN) and Swin Transformer has been shown effective at both extracting contextual features and representing global long-range dependency features in detection and segmentation tasks [17, 18, 19].

In this paper, we propose an improved small object detection model called CST-YOLO, by leveraging both YOLOv7 and Swin Transformer [16] for blood cell detection. To our knowledge, CST-YOLO is the first object detector of Transformer with YOLOv7. The proposed CST-YOLO introduces several novel features in the YOLOv7 as follows:

1) We design a convolutional or CNN-Swin Transformer (CST) module based on Swin Transformer in the backbone and introduce it into the feature extraction module of YOLOv7, which can enhance the receptive field of the model and extract the target feature information better.
2) We develop an adaptive feature fusion module called Weighted ELAN (W-ELAN), which can dynamically fuse the feature map data with the effective feature map and restrain the invalid feature map.
3) We apply a Multiscale Channel Split (MCS) module to extract feature information from the input feature map with different receptive fields.
4) We build a Concatenation Convolutional Layers (CatConv) module to enhance the feature fusion capability and achieve better target detection by fusing more feature data.

Experimental evaluation on three blood cell datasets shows superior detection performance of CST-YOLO over advanced DEtection TRansformer (DETR) and YOLO detectors.

## 2. METHODS

The proposed CST-YOLO model is shown in Fig. 1. It incorporates four new components in the YOLOv7, i.e., CST, W-ELAN, MCS, and CatConv.

### 2.1. CNN-Swin Transformer

To better capture global information in images, we designed a novel CNN-Swin Transformer module in a parallel fusion structure (Fig. 2) distinct from the existing ones [17, 18, 20, 19] and introduced it into YOLOv7 architecture. The CST firstly uses two parallel $1 \times 1$

This work is supported by the Monash University Malaysia and the Ministry of Higher Education, Malaysia under Fundamental Research Grant Scheme FRGS/1/2023/ICT02/MUSM/02/1.

*Corresponding author, e-mail: ting.cheeming@monash.edu.

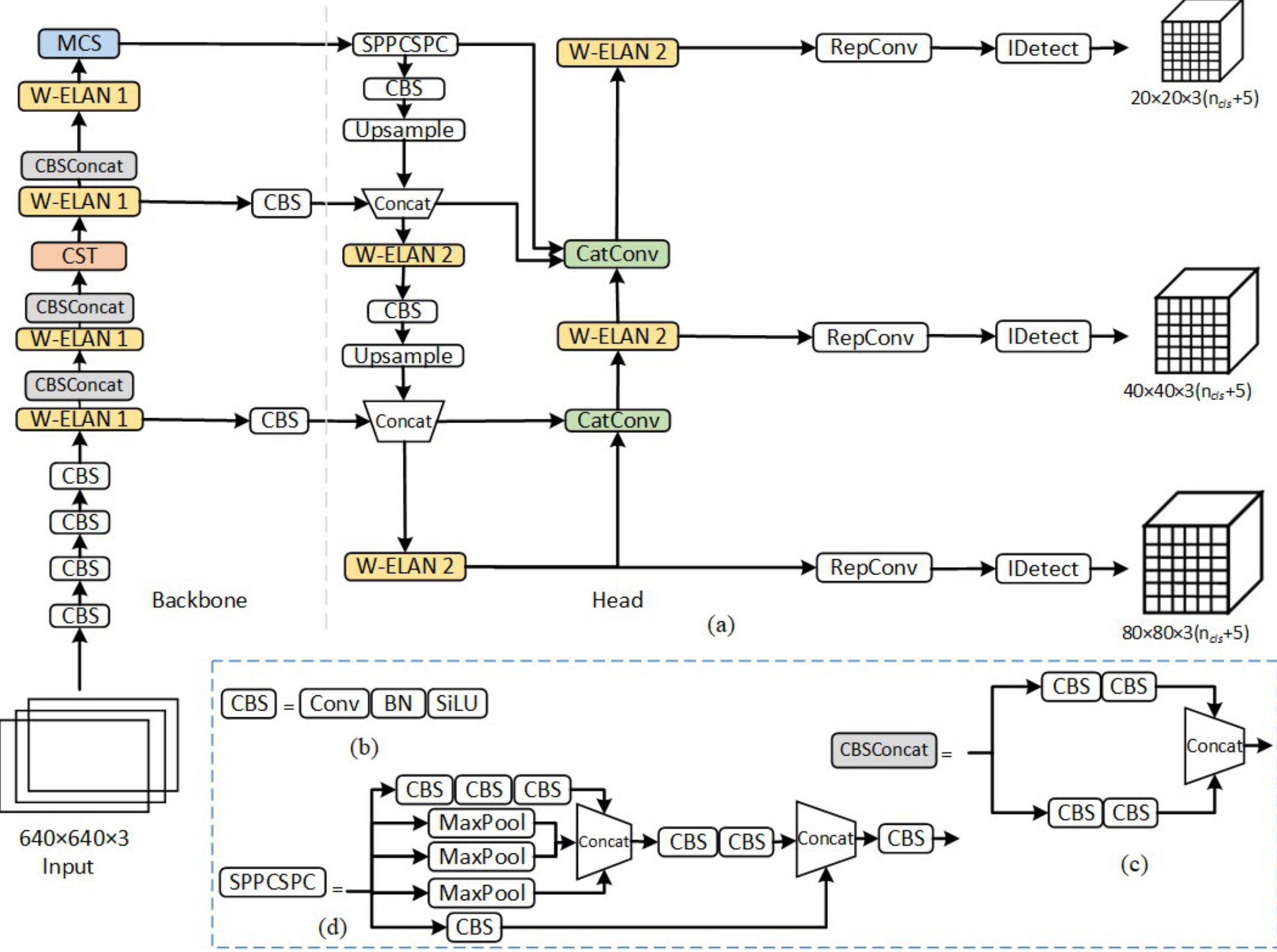

**Fig. 1**. (a) Overview of CST-YOLO. The architecture of CST-YOLO is based on YOLOv7 and incorporates new modules colored: CST, W-ELAN, MCS, and CatConv. The first CBS on the left of (c) CBSConcat replaces max pooling (MaxPool) in MPConv of YOLOv7. (b) CBS, (d) SPPCSPC, RepConv, and IDetect are existing modules in YOLOv7.

convolutions to adjust the number of channels of the input feature map to obtain two outputs, one of which is used as input to the Swin Transformer, which captures the global feature information of the input feature map and then splices the two outputs. The CST module can extract features from the input feature map with different receptivity, which greatly improves the model's ability to represent the input feature map, thus enhancing the model's detection performance. Compared to the existing approaches of the CNN-Swin Transformer fusion cited in the paper, our proposed CST module has a novel fusion structure, which concatenates two sets of combined features in a parallel structure of CNN and Swin Transformer after a series connection of them. It can enhance the receptive field of the model and extract the small object contextual information better with a window-shifted self-attention mechanism.

The Swin Transformer consists of two layers of self-attention mechanisms, using a normal window and shift window, respectively. At the former layer, the input feature map is first divided into windows (Fig. 3 (a)), and then the self-attention mechanism is used for each window without any information interaction between the windows. This mechanism improves the detection speed but reduces the receptive field of the model, which is not conducive to feature extraction in the network. The shift window self-attention mechanism (Fig. 3 (b)) is subsequently introduced, where the window division of the normal window self-attention mechanism is first window-shifted, and the information in one window after the shift

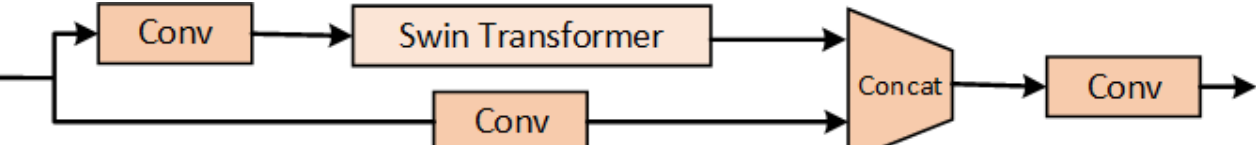

**Fig. 2**. The architecture of CNN-Swin Transformer module.

is equivalent to fusing multiple windows in the normal window self-attention mechanism.

For the blood cell image example in Fig. 3, the input of the window in the middle blue area in layer $i$+1 comes from the same area in layer $i$, while the blue area in layer $i$+1 contains the information of four windows in layer $i$. That is, each window in layer $i$+1 is equivalent to the information of four windows in layer $i$. This overcomes the problem that information can not be exchanged between different windows. By using both $i$-th and $i$+1th layers in conjunction, it reduces the number of parameters as required in the traditional self-attention mechanism, and also alleviates the problem of the reduced receptive field of the normal window self-attention mechanism. By using the $i$-th and $i$+1th layers in conjunction, not only can the problem of an excessive number of parameters in the traditional self-attention mechanism be solved, but also the problem that the receptive field of the normal window self-attention mechanism is reduced.

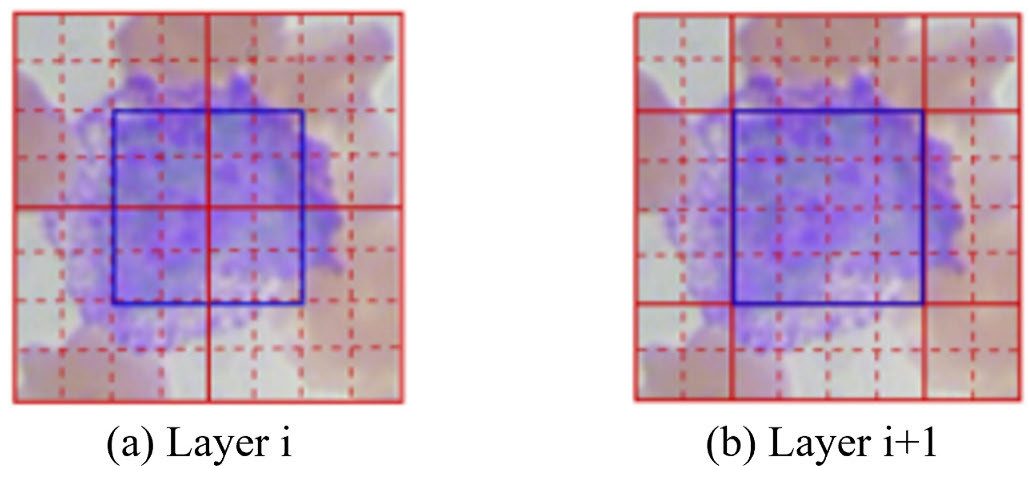

**Fig. 3**. The shift window of Swin Transformer.

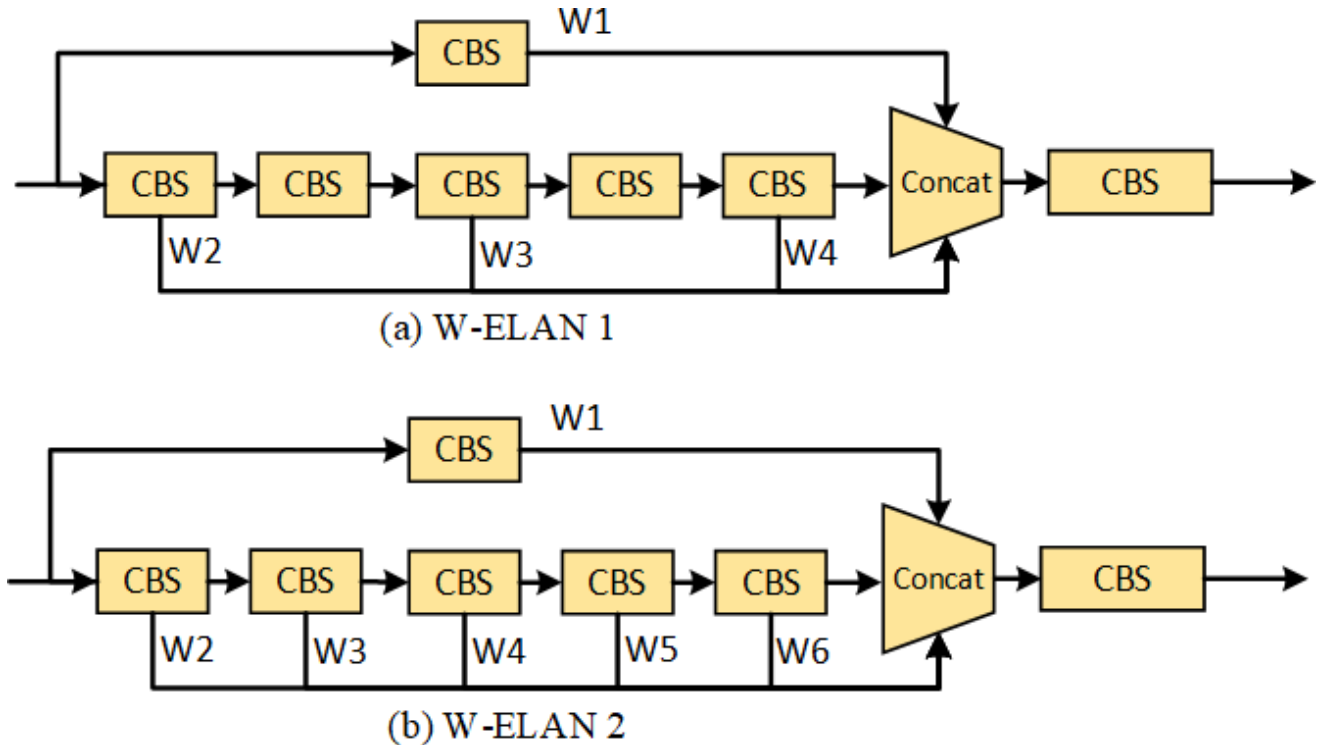

**Fig. 4**. The architectures of two variants of the Weighted ELAN module. (a) W-ELAN 1. (b) W-ELAN 2.

### 2.2. Weighted ELAN

Inspired by weighted feature fusion in EfficientDet [21], we develop a Weighted ELAN, a new variant different from E-ELAN [6, 22], to facilitate dynamic feature fusion. The architecture of W-ELAN is given in Fig. 4. A weight is applied to each feature map during the feature map splicing process, and it is a learnable parameter in the training process.

In order to prevent the weights $w_i$ corresponding to different feature maps from varying greatly, the learned weights have to be normalized as follows

$$w_i = \frac{w_i'}{\sum_{j=0} w_j' + \xi} \qquad (1)$$

where $w_i$ is the normalized feature map weight, $w_i'$ is the weight of the network output during training, and $\xi$ is a fixed small value that prevents the denominator from being 0. During the training process, the network automatically assigns a larger weight to a feature if it makes a larger contribution to the final network's prediction, while a smaller weight is assigned to a feature with a smaller contribution to the final network's prediction. We noticed that this process is entirely determined by the training process without active jamming.

### 2.3. Multiscale Channel Split

Many studies have shown that significant differences exist in the extraction of target feature information in computer vision tasks with different settings of receptive field size in convolution. The smaller the receptive field is, the more localized image information can be observed in the network. Instead, the larger the receptive field is, the better the network is at understanding the global information of the input features. Drawing the idea of channel split in ShuffleNet V2 [23], we apply an MCS module (Fig. 5) to improve the model's ability to perceive target feature information at different scales.

The module passes the input feature maps through four different kernel average pooling layers to obtain feature maps of different sizes, then performs channel adjustment using $1 \times 1$ convolution to obtain four feature maps, where all channel numbers are 256. To concatenate the four feature maps, we first apply four up-sampling operations with different up-sampling coefficients to adjust the width and height of the four feature maps to a uniform size and then perform concatenation. Suppose the input feature map size is (B, C, W, H), after the above operations, the output feature map is (B, 1000, W, H). This feature map is then passed through the Average Pooling (AvgPool) where the average value of all the pixels in the batch is selected to obtain a feature map of size (B, 1000, 1, 1), and after the Sigmoid activation function, the input feature values are normalized to between 0 and 1, which will be multiplied with the input feature map on the channel. This is equivalent to applying different attention to each channel of the input features, allowing the model to dynamically filter the channels with more feature information and suppress those with less.

To reduce the parameters of the MCS module, the feature map obtained after the above operations is divided into four along the channel direction, after which the four feature maps are summed, and the resulting feature maps are adjusted by using $1 \times 1$ convolution to the channel. A residual connection is also added to obtain the final output of the module.

### 2.4. Concatenate Convolutional Layers

To enhance the feature fusion capability of the model, we modify the MPConv structure of the feature fusion part in Yolov7 by first replacing the max pooling (MaxPool) where the maximum pixel value of the batch is selected, hence producing CBSConcat in the backbone. Then, to extract more feature information using convolution, we multiply the relevant features from the downsampling part by 2 as the input to a Concatenate Convolutional Layers module (Fig. 6) in the neck. At the same time, to ensure the uni-size of the feature map, we adjust the size with a convolution whose step is 2 before feature fusion. In addition, we use RepConv + IDetect in YOLOv7 as the detection head. Essentially, RepConv is RepVGG [24] without identity connection and the latter is adopted in the reparameterizable backbone of YOLOv6 [25] and YOLOv6 v3.0 [26] at training.

## 3. EXPERIMENTS

### 3.1. Datasets

We evaluated the proposed CST-YOLO on three blood cell detection datasets, including Blood Cell Count and Detection (BCCD) [27], Complete Blood Count (CBC) [28, 29], and Blood Cell Detection (BCD) [30] datasets. These are small-scale object datasets with annotations of three classes of blood cells (RBCs, WBCs, and platelets. The three datasets differ in quantity, resolution, and division.

BCCD Dataset contains 364 images with $640 \times 480$ pixels. CBC dataset contains 360 blood smear images and each image is resized to $640 \times 480$ resolution. There are 364 images with $416 \times 416$ pixels resolution in the BCD dataset. The data partitioning of the three datasets into training, testing, and validation sets is given in Table 1. For CBC and BCD, we followed the original data partitioning as suggested in the sources of the datasets.

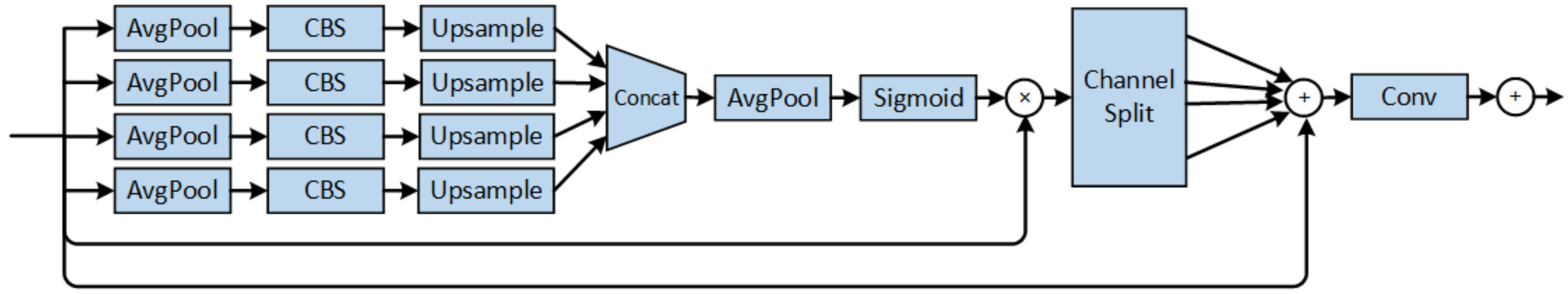

**Fig. 5**. The architecture of Multiscale Channel Split module.

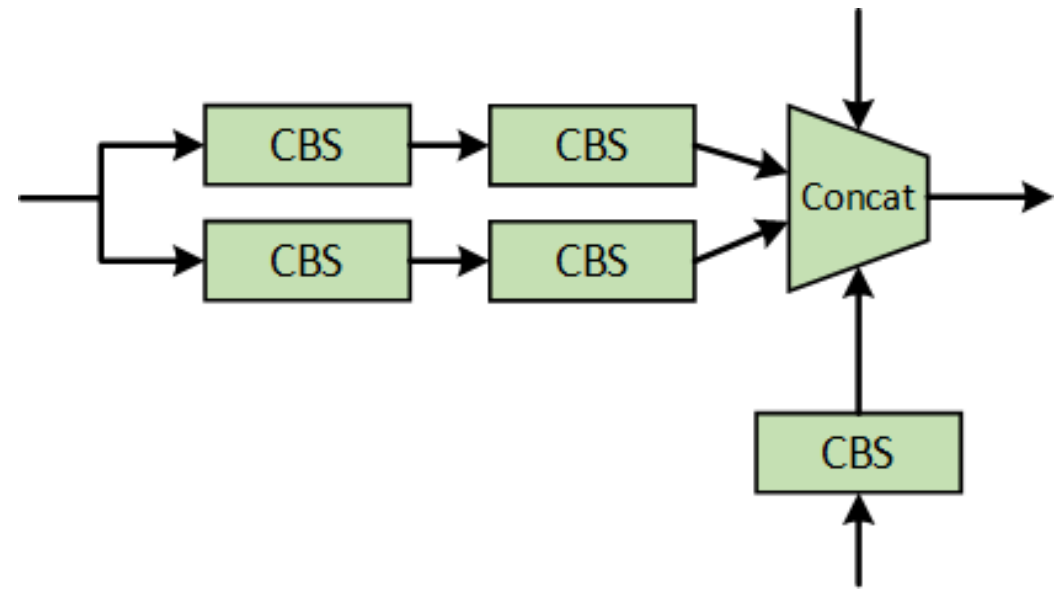

**Fig. 6**. The Concatenate Convolutional Layers module.

**Table 1**. Number of examples in BCCD, CBC, and BCD.

| Dataset | Training | Validation | Testing | Total |
|---|---|---|---|---|
| BCCD | 205 | 87 | 72 | 364 |
| CBC | 300 | 0 | 60 | 360 |
| BCD | 255 | 73 | 36 | 364 |

**Table 2**. Performance comparison of RT-DETR, YOLOv5x, YOLOv7 and CST-YOLO for blood cell detection. Results are mAP@0.5 for each blood cell type and overall performance. The original code of all RT-DETR versions only prints results for all classes.

| Dataset | Model | RBCs | WBCs | Platelets | Overall |
|---|---|---|---|---|---|
| BCCD | RT-DETR [31] | – | – | – | 0.875 |
| | YOLOv5x [7] | **0.877** | 0.977 | 0.915 | 0.923 |
| | YOLOv7 [6] | 0.829 | 0.977 | 0.883 | 0.896 |
| | **CST-YOLO** | 0.869 | **0.984** | **0.928** | **0.927** |
| CBC | RT-DETR [31] | – | – | – | 0.855 |
| | YOLOv5x [7] | 0.930 | 0.995 | **0.942** | 0.955 |
| | YOLOv7 [6] | 0.917 | 0.995 | 0.912 | 0.941 |
| | **CST-YOLO** | **0.947** | 0.995 | 0.927 | **0.956** |
| BCD | RT-DETR [31] | – | – | – | 0.784 |
| | YOLOv5x [7] | 0.857 | 0.820 | 0.975 | 0.884 |
| | YOLOv7 [6] | 0.785 | 0.874 | 0.974 | 0.878 |
| | **CST-YOLO** | 0.857 | **0.899** | **0.978** | **0.911** |

### 3.2. Implementation Details

The CST-YOLO was trained with NVIDIA® Telsa® V100 and tested with NVIDIA® GeForce RTX® 2060. The hyperparameters used in the training are a learning rate of 0.001 with a cosine annealing learning rate schedule and a weight decay rate of 0.0005. The training parameter batch size is set to 20, and the epoch is 150 in the learning phase.

### 3.3. Results

To evaluate object detection performance, we use average precision (AP) as metrics for each type of detected blood cell and mean average precision (mAP) @0.5 (averages over all types) as the metric of the overall model. Table 2 shows the APs for different types of blood cells and mAP@0.5 using different detectors. We compare the proposed CST-YOLO with both non-YOLO and YOLO models: RT-DETR [31] with ResNet50 backbone (i.e., rtdetr_r50vd in the RT-DETR model zoo) which is a Transformer-based non-YOLO detector, YOLOv5x v6.1 (no relevant updates on detection task in the newer versions of v6.2 and v7.0, only classification and segmentation checkpoints added in v6.2 and v7.0 respectively), and YOLOv7. Results show that the CST-YOLO overall mAP@0.5 improved by 3.1%, 1.5%, and 3.7%, respectively, compared to the YOLOv7 results. AP values of the RBCs, WBCs, and platelets have all some improvement to YOLOv7. CST-YOLO with 47.5M parameters even surpasses the extra large object detector YOLOv5x (86.7M parameters), which is twice larger than YOLOv7 (36.9M parameters) on the overall mAP.

The potential limitation of the proposed CST-YOLO is mainly on increased computational complexity. The computational performance of CST-YOLO is 235.6 giga floating-point operations (GFLOPs), which is bigger than YOLOv7 with 103.4 GFLOPs. However, the CST-YOLO gives a substantial improvement in accuracy despite a slight increase in computational effort.

### 3.4. Ablation Study

To validate which of the proposed modules plays a vital role in improving the accuracy of the BCCD dataset, we investigate the effect on the performance of the proposed modules. We separately restore the original modules of YOLOv7 which are replaced by CST, W-ELAN, MCS, and CatConv & CBSConcat (i.e., w/ MaxPool).

In Table 3, it is seen that the overall mAP@0.5 of all methods that remove one of the proposed modules both decrease than CST-YOLO, so each proposed module brings forth a positive effect on the improvement of CST-YOLO. The method without the CST module has the biggest falls in the mAP@0.5, which demonstrates the importance of the CST module is more crucial than others of our proposed modules in CST-YOLO. As platelets are the smallest and most indistinct objects in the images, another point of our analysis focuses on them. From decreasing values of APs on platelets, we judge the CST module has a strong effect on the accuracy improvement of small blood cell detection, which is the key point influencing overall performance.

**Table 3**. Abation study of the proposed modules. Results are mAP@0.5 for each blood cell type and overall performance

| Method | RBCs | WBCs | Platelets | Overall |
|---|---|---|---|---|
| w/o CST | 0.841 | 0.849 | 0.900 | 0.894 |
| w/o W-ELAN | 0.860 | 0.866 | 0.987 | 0.904 |
| w/o MCS | 1.000 | 0.762 | 0.987 | 0.910 |
| w/ MaxPool | 0.806 | 0.856 | 0.901 | 0.899 |

**Table 4**. Performance comparison of YOLOv7 and CST-YOLO for all classes of the TinyPerson dataset. The best results are shown in bold.

| Model | Precision | Recall | mAP@0.5 | mAP@0.5:0.95 |
|---|---|---|---|---|
| YOLOv7 | 0.530 | **0.361** | 0.335 | 0.111 |
| **CST-YOLO** | **0.565** | 0.358 | **0.344** | 0.111 |

### 3.5. Generalizability Study

To demonstrate the generalizability of our proposed CST-YOLO, we conducted independent experiments on the TinyPerson dataset [32] where the size of most annotated objects is less than $20 \times 20$ pixels in line with the small object detection setting in this paper. The TinyPerson dataset contains 1,610 images with 72,651 annotated human objects with resolution visual effects. The train and test sets of this dataset have been divided for the Tiny Object Detection Challenge of the ECCV 2020 workshop.

Table 4 compares the performance between the original YOLOv7 and CST-YOLO on the TinyPerson dataset in different domains from medical images for external validation. The experimental results demonstrate the generalizability effectivenss of CST-YOLO in small object detection on both medical and natural images.

### 3.6. Visual Illustration

Some heat maps are generated by Gradient-weighted Class Activation Mapping (Grad-CAM) [33] from the BCCD dataset. Examples of heat maps and a special case are shown in Fig. 7 and Fig. 8, respectively. From the visual explanations from deep networks via gradient-based localization of the sample BloodImage_00261.jpg in Fig. 7, we have a much more precise region of emphasis that locates the different types of blood cells and know that the model classifies this input image due to its intrinsic features, not a general region in the image. Fig. 8 is a special case of the sample BloodImage_00340.jpg because no Grad-CAM heat map is solely generated for platelets in this case. There are platelets in the image, but they are not annotated in the file BloodImage_00340.xml of this sample. From this case, we conclude that unsupervised or semi-supervised learning's ability to detect objects from unlabeled data with the YOLO framework needs to be further studied in future work.

## 4. CONCLUSION

We develop a novel method, CST-YOLO, for small-scale object detection by incorporating a CNN-Swin Transformer module in an improved YOLOv7-like architecture. Evaluated on blood cell detection on three benchmarking datasets, the proposed model provides substantially better detection precision than YOLOv5 and YOLOv7. This suggests that the incorporation of a CNN-Transformer fusion can potentially increase detection performance for small objects.

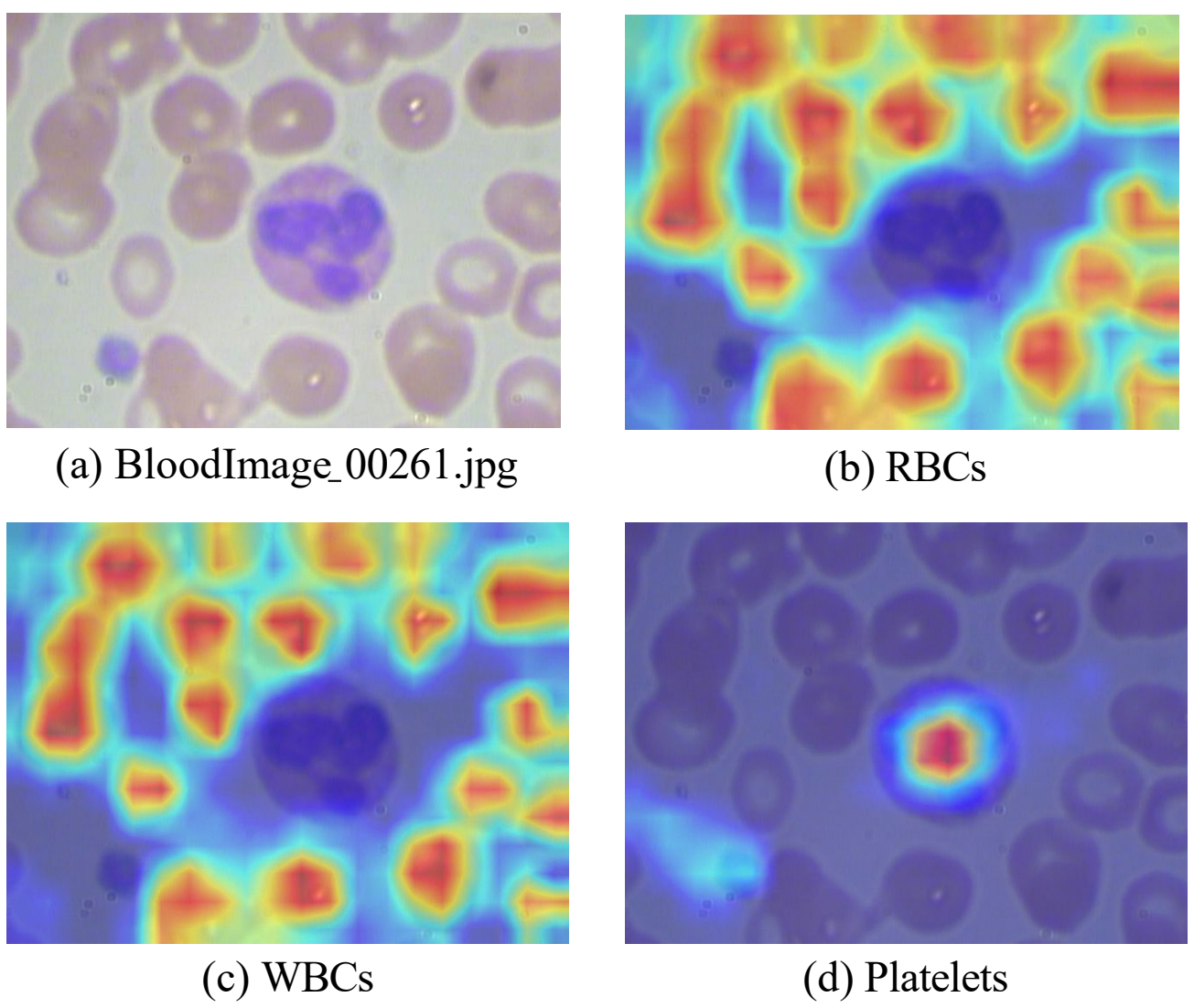

(a) BloodImage_00261.jpg (b) RBCs (c) WBCs (d) Platelets

**Fig. 7**. Example of Grad-CAM heat maps. Input the original blood cell image (a) BloodImage_00261.jpg to be detected and generate heat maps. The three heat maps with gradient-weighted class activation mapping respectively emphasize the objects in different classes: (b) RBCs, (c) WBCs, and (d) Platelets, in the blood cell image, and de-emphasize the other two classes.

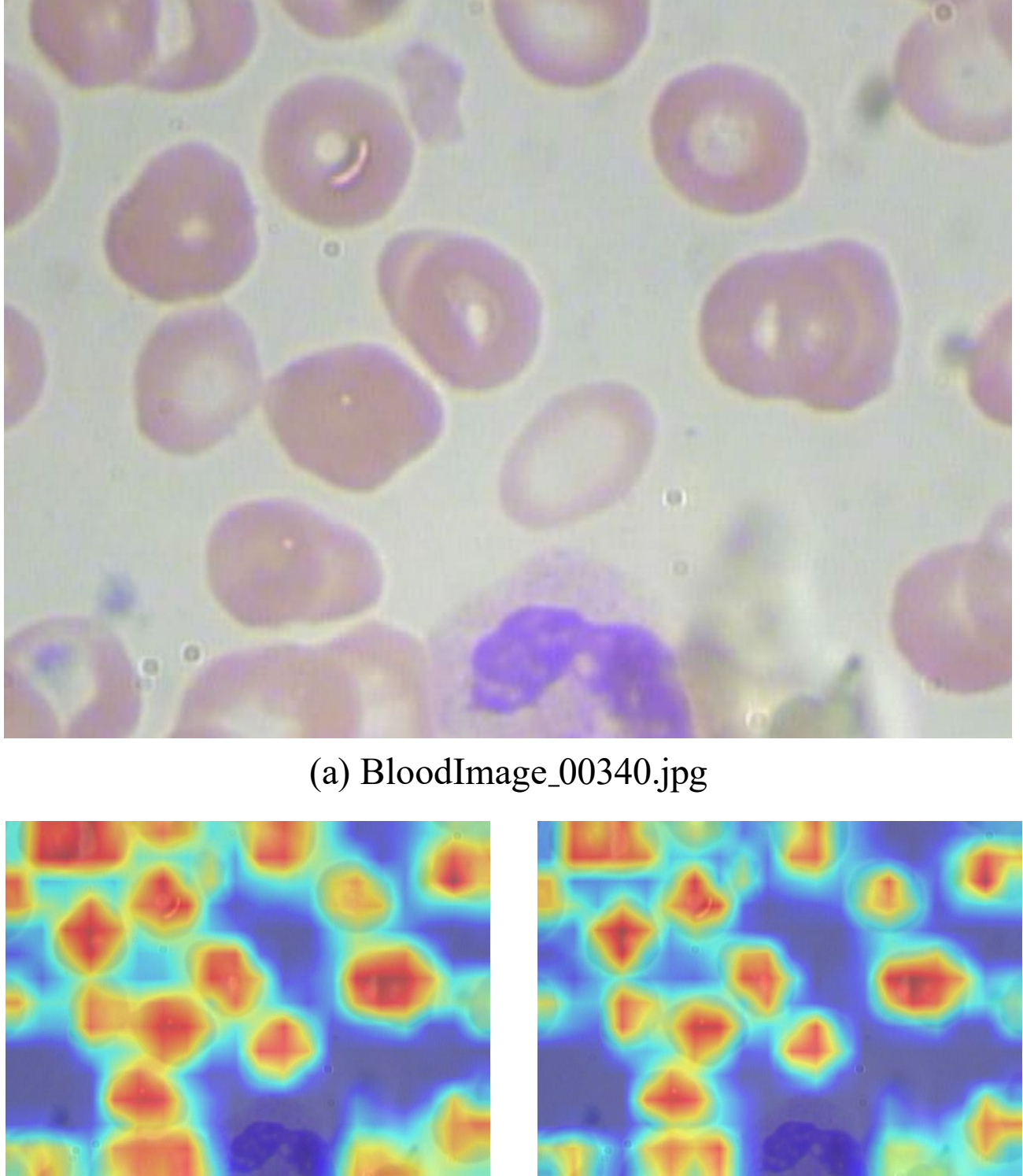

(a) BloodImage_00340.jpg (b) RBCs (c) WBCs

**Fig. 8**. Example of a special case. The heat maps of (b) RBCs, and (c) WBCs are generated by inputting the original blood cell image (a) BloodImage_00340.jpg.